\documentclass[nopreprintline,12pt]{elsarticle}
\DeclareUnicodeCharacter{2061}{}
\DeclareUnicodeCharacter{03C3}{}
\usepackage{amssymb}
\usepackage{xcolor}
\usepackage{amsmath}
\usepackage{multirow}
\usepackage{multicol}
\usepackage{diagbox}
\usepackage{booktabs}
\usepackage{wrapfig}
\usepackage{picinpar}
\usepackage{graphicx}
\usepackage{subfigure}
\usepackage{CJKutf8}
\usepackage{comment}

\newcommand{\xu}[1]{\textcolor{blue}{#1}}

\newcommand{\MR}[1]{\textcolor[rgb]{0,0,0} {#1}}
\newcommand{\MiR}[1]{\textcolor[rgb]{0,0,0} {#1}}

\journal{Artificial Intelligence}

\begin{document}
\begin{CJK}{UTF8}{gbsn}
\begin{frontmatter}



\title{Saliency-Aware Regularized Graph Neural Network}


\author[label1]{Wenjie Pei\corref{cor1}} 
\author[label2]{WeiNa Xu\corref{cor1}}
\author[label3]{Zongze Wu}
\author[label4]{Weichao Li}
\author[label2]{Jinfan Wang}
\author[label1]{Guangming Lu}
\author[label3]{Xiangrong Wang\corref{cor2}}

\cortext[cor1]{Equal contribution.}
\cortext[cor2]{Corresponding author.}

\affiliation[label1]{organization={Department of Computer Science, Harbin Institute of Technology at Shenzhen},
            city={Shenzhen},
            postcode={518172}, 
            country={China}}

\affiliation[label2]{organization={Institute of Future Networks, Southern University of Science and Technology},
            city={Shenzhen},
            postcode={518055}, 
            country={China}}

\affiliation[label3]{organization={College of Mechatronics and Control Engineering, Shenzhen University},
            city={Shenzhen},
            postcode={518060}, 
            country={China}}

\affiliation[label4]{organization={Peng Cheng Laboratory},
            city={Shenzhen},
            postcode={518066}, 
            country={China}}

\begin{abstract}

The crux of graph classification lies in the effective representation learning for the entire graph. Typical graph neural networks focus on modeling the local dependencies when aggregating features of neighboring nodes, and obtain the representation for the entire graph by aggregating node features. Such methods have two potential limitations: 1) the global node saliency w.r.t. graph classification is not explicitly modeled, which is crucial since different nodes may have different semantic relevance to graph classification; 2) the graph representation directly aggregated from node features may have limited effectiveness to reflect graph-level information. In this work,  we propose the Saliency-Aware Regularized Graph Neural Network (\emph{SAR-GNN}) for graph classification, which consists of two core modules: 1) a traditional graph neural network serving as the backbone for learning node features and  2) the Graph Neural Memory designed to distill a compact graph representation from node features of the backbone. We first estimate the global node saliency by measuring the semantic similarity between the compact graph representation and node features. Then the learned saliency distribution is leveraged to regularize the neighborhood aggregation of the backbone, which facilitates the message passing of features for salient nodes and suppresses the less relevant nodes. Thus, our model can learn more effective graph representation. We demonstrate the merits of \emph{SAR-GNN} by extensive experiments on seven datasets across various types of graph data. Code will be released.
\end{abstract}




\end{frontmatter}


\section{Introduction}
\label{sec:intro}

Successful graph classification requires effective representation learning not only for each node but also for the entire graph. Existing methods based on graph neural networks~\cite{scarselli2008graph, micheli2009neural} have achieved significant progress in node representation learning, benefiting from the excellent feature refinement through neighborhood aggregation for each node.  Most of these methods focus on modeling the local dependencies between neighboring nodes when performing neighborhood aggregation. Prominent examples include GCN~\cite{GCN} which aggregates neighborhood features by considering the node degree distribution, GraphSAGE~\cite{graphsage} as well as GIN~\cite{xu2018powerful} which seek to learn effective aggregating functions, and GAT~\cite{GAT} which employs an attention mechanism to learn aggregating weights. While these methods are able to learn effective node representations in a graph, they ignore the global saliency of each node, \MR{formulated as the semantic similarities between features of each node and the graph representation in a latent space}. Nevertheless, it is crucial to perceive global node saliency for graph classification since different nodes potentially have different relevance. Noisy or irrelevant nodes may have adverse effects on graph classification. 

A straightforward way of modeling such global saliency is to employ attention mechanism to calculate the attention distribution, 
as SAGPool~\cite{lee2019self} and ChebyGIN~\cite{knyazev2019understanding} behave. \MR{However, such implicit modeling way has limited effectiveness since it estimates the attention scores solely based on local node features and lacks the explicit modeling of semantic similarities between each node and the global graph representation.} 
Other typical methods of modeling global node saliency include Graphormer~\cite{ying2021transformers} interpreting node degrees as centrality, DGM~\cite{DGM} utilizing node connectivities to compute a lens function, and DGCNN~\cite{DGCNN} equating the graph convolution output with continuous Weisfeiler-Lehman (WL)~\cite{leman1968reduction} for sorting nodes (termed `SortPooling'). \MR{Nevertheless, all these methods still suffer from the limitation that they do not explicitly measure the semantic similarities between each node and the global graph representation, which essentially characterizes the saliency.}

Another essential concern of graph classification is how to learn the effective representation for the entire graph. A commonly adopted approach by most existing methods based on graph neural networks~\cite{ECC, diffpool, DGCNN, eigenpooling, gao2019graph}, is to first learn node features and then aggregate all node features to produce the graph representation via a readout function, which is typically implemented by various pooling operations. Despite the simplicity of such approach, the obtained graph representation may have limited effectiveness. This is because 
most model capacity is allocated for learning node features rather than the graph representation, and the quality of obtained graph representation depends heavily on that of the node features.

To address the above two potential limitations, in this work we propose the Saliency-Aware Regularized Graph Neural Network (\emph{SAR-GNN}) for graph classification which is built upon traditional graph neural networks. It first measures the saliency of each node in a global view, \MR{namely the semantic similarities between each node and the graph representation}, and then leverages the learned saliency distribution to regularize the graph neural network. Specifically, the \emph{SAR-GNN} consists of two core modules: 1) a traditional graph neural network serving as the backbone for learning node features and 2) Graph Neural Memory which is designed to distill a compact graph representation for the entire graph from node features of the backbone. Both the backbone network and the Graph Neural Memory are iteratively stacked to refine node features and the graph representation progressively. The learned compact graph representation is used to estimate the global saliency of each node by measuring the semantic similarities between the graph representation and node features. Then the obtained saliency distribution is used to regularize the aggregating weights of the backbone network when performing neighborhood aggregation for each node, which facilitates the message passing of features for the salient nodes while suppressing the feature propagation for the less relevant nodes to graph classification. As a result, the Graph Neural Memory can distill more relevant features to graph classification from the salient nodes when learning the compact graph representation. Two modules work interdependently with interactions between each other in each stacked layer. The obtained compact graph representation is finally utilized for graph classification. 


\MR{To conclude, we highlight following contributions.}
\begin{itemize}
    \item \MR{We propose the Saliency-Aware Regularized Graph Neural Network (\emph{SAR-GNN}), a novel framework for graph classification, which consists of two core modules: a backbone network for learning node features and the Graph Neural Memory for distilling the compact graph representation. Two modules are optimized interdependently with interactions between each other in each stacked layer and thus both the node features and the graph representation can be refined iteratively. Such framework enables the proposed \emph{SAR-GNN} to model the global node saliencies in an explicit manner by measuring the semantic similarities between each node and the graph representation.}
    \item \MR{We design an effective saliency-aware regularization mechanism, which utilizes the learned global node saliencies to regularize the feature aggregation for each node in such a way that it facilitates the message passing of features for the salient nodes while suppressing the feature propagation for the less relevant nodes to graph classification. As a result, the Graph Neural Memory can distill more relevant features to graph classification from the salient nodes and thereby produce more effective graph representation for graph classification.}
    \item \MR{The proposed \emph{SAR-GNN} can be readily applied to most of the existing graph neural networks which perform neighborhood aggregation for refining node features. In particular, we instantiate the backbone of our model with four classical types of graph neural networks, namely GCN, GraphSAGE, GIN and GAT respectively. Then we conduct extensive experiments both quantitatively and qualitatively on seven challenging datasets across various types of graph data, which demonstrate 1) the effectiveness of our method via thorough ablation studies and complexity analysis, and 2) the favorable performance of our model compared to the state-of-the-art methods for graph classification.}
\end{itemize}
\section{Related Work}

Traditional approaches to graph classification, preceding the popularity of graph neural networks, typically employ graph kernel functions to measure the similarity between pairs of graphs~\cite{wu2020comprehensive, shervashidze2011weisfeiler, zhang2021deep}. Our \emph{SAR-GNN} is built upon graph neural networks and is designed to: 1) regularize the neighborhood aggregation of graph neural networks 
and 2) learn effective representation for the entire graph. Thus, we discuss related work to our method from these two perspectives below.

\noindent\textbf{Neighborhood aggregation of Graph Neural Networks.}
The core idea shared across various graph neural networks is to perform neighborhood aggregation for each graph node to refine the node features iteratively~\cite{chen2019powerful, Errica2020A}. 
Typical convolutional neural networks~\cite{alexnet} are successfully adapted to graph data either in the spectral domain~\cite{bruna2013spectral, defferrard2016convolutional} or spatial domain~\cite{micheli2009neural, niepert2016learning}. 
In particular, GCN~\cite{GCN} presents a scalable and efficient implementation of graph convolution. GraphSAGE~\cite{graphsage} explores multiple potential aggregating functions while Graph Isomorphism Network (GIN)~\cite{xu2018powerful} builds upon GraphSAGE and presents a framework which is as theoretically powerful as the Weisfeiler- Lehman graph isomorphism test~\cite{leman1968reduction}. Another research line of modeling neighborhood aggregation is to learn the aggregating weights directly per edge around the center node. Edge-Conditioned Convolution (ECC)~\cite{ECC} learns aggregating weights conditioned on edge labels while GAT~\cite{GAT} employs an attention mechanism to measure compatibility between neighboring nodes for each edge.

All aforementioned methods focus on modeling the local dependencies between neighboring nodes when performing feature aggregation whilst the relevance of each node to the task of graph classification is ignored. \MR{While the node saliency can be modeled by the attention mechanism roughly, 
as ChebyGIN~\cite{knyazev2019understanding} and SAGPool~\cite{lee2019self} do, this type of methods has limited effectiveness in that it solely relies on the attention model without explicit modeling of saliency.} 
Besides, the global node saliency has been also modeled based on 1) node degrees by Graphormer~\cite{ying2021transformers}, or 2) the connectivities between nodes by DGM~\cite{DGM}, or 3) the graph convolution output by DGCNN~\cite{DGCNN}. Nevertheless, all these methods do not explicitly model the saliency yet, namely the semantic similarties between each node and the global graph representation. In contrast, our \emph{SAR-GNN} measures the node saliency by modeling the compatibility between the learned graph representation and the node features. The learned node saliency is then leveraged to regularize the modeling of neighborhood aggregation.

\noindent\textbf{Representation learning for the entire graph.} Most of the existing methods for graph classification learn the representation for the entire graph in an indirect manner: they allocate most of the model capacity to learning effective node features, and obtain the representation for the entire graph by simply aggregating node features via a readout function, mostly implemented as a variety of pooling functions. Typical examples include DiffPool~\cite{diffpool} which designs an adaptive pooling method that collapses nodes hierarchically, ECC which pre-calculates a pooling map to coarsen graphs, DGCNN~\cite{DGCNN} that sorts nodes first using SortPool before pooling,  EigenPooling~\cite{eigenpooling} which is proposed based on graph Fourier transform to preserve graph structure during pooling, Graph U-Nets~\cite{gao2019graph} which performs top-K pooling scheme, and SAGPool~\cite{lee2019self} which identifies the important nodes using the self-attention mechanism~\cite{transformer}. On the other hand, ESAN~\cite{ESAN} represents a large graph as a set of distinguishable subgraphs based on the predefined policy. Unlike aforementioned prior work, we design a module termed as Graph Neural Memory specifically to distill a compact graph representation from node features progressively using a cross-attention mechanism. In particular, the compact graph representation and the node features are refined by two modules interdependently.

\MiR{Although our model is mainly designed for effective graph representation learning, another potentially advantageous application of it, which is worth noting, is that the derived node saliency distribution by our \emph{SAR-GNN} can be naturally used for interpretable explanation of GNN predictions. In this sense, our model is essentially consistent with GNNExplainer~\cite{GNNExp} which identifies compact subgraphs crucial for GNN's predictions for interpretation, as well as PGExplainer~\cite{PGExp} that is a generalizable GNN explainer by parameterizing the generation process of explanations.}
\section{Saliency-Aware Regularized Graph Neural Network}
Built upon traditional graph neural networks, our \emph{SAR-GNN} first captures the global node saliency w.r.t. the task of graph classification. The learned node saliency is then leveraged to regularize the neighborhood aggregation of graph neural networks and facilitate the massage passing of features for salient nodes, thus our model can learn more effective representations for the entire graph. 

\subsection{Overview}
\begin{figure*}[!t]
\centering
\includegraphics[width=1.0\linewidth]{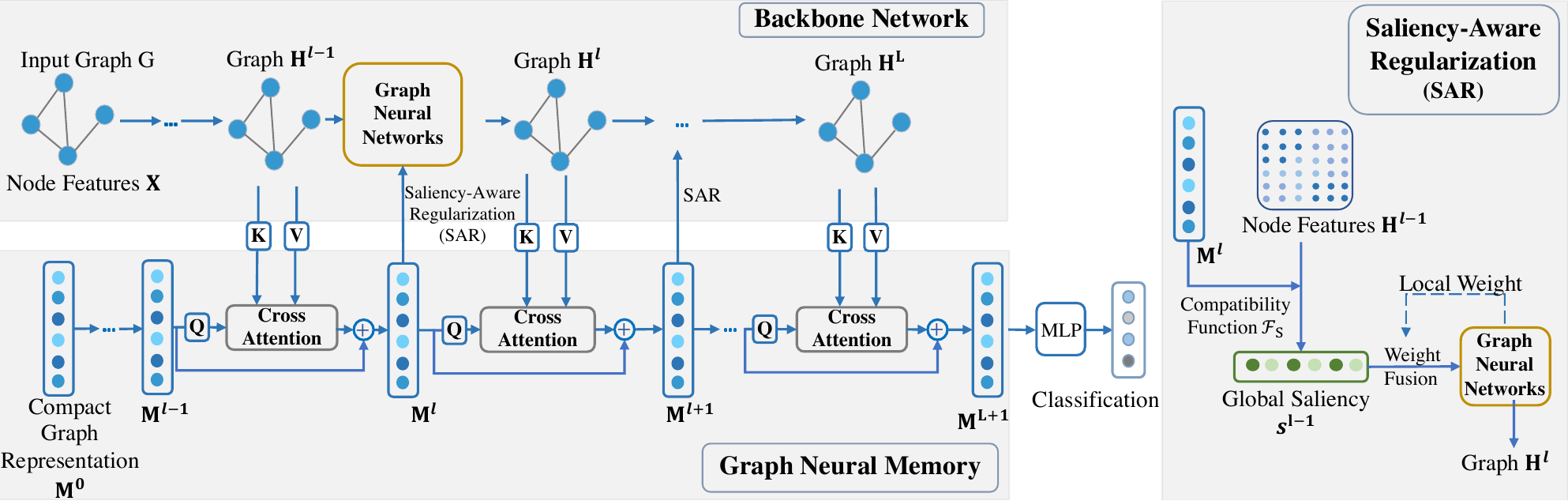}
\caption{Architecture of the Saliency-Aware Regularized Graph Neural Network (SAR-GNN). It consists of two core modules: 1) a traditional graph neural network serving as the backbone network for learning node features and 2) the Graph Neural Memory for distilling a compact graph representation from node features of the backbone. The learned graph representation is leveraged to measure the global node saliency and regularize the backbone. Thus, two modules work interdependently to refine node features and the compact graph representation in an iterative manner. The operation of Saliency-Aware Regularization in the left panel is elaborated in the right panel.}
\label{fig:architecture}
\end{figure*}

Figure~\ref{fig:architecture} illustrates the architecture of our \emph{SAR-GNN}, which consists of two core modules: 1) a traditional graph neural network serving as the backbone for learning features of each node in a graph and 2) Graph Neural Memory which is specifically designed for distilling a compact graph representation for the entire graph from the node features of the backbone. Two modules work interdependently to refine both the compact graph representation and node features progressively through the stacked layers.

Taken a graph $G$ with $N$ nodes as input, \emph{SAR-GNN} first employs the Graph Neural Memory to perform feature distillation from the node features in the backbone network and obtains a compact graph representation for the whole graph:
\begin{equation}
    \mathbf{M}^l = \mathcal{F}_\text{M}(\mathbf{H}^{l-1},\mathbf{M}^{l-1}).
    \label{eqn:fm}
\end{equation}
Here $\mathbf{M}^l \in \mathbb{R}^{d_M}$ denotes the compact graph representation with $d_M$ dimensions at the $l$-th stacked layer, which is refined iteratively from the previous representation $\mathbf{M}^{l-1}$ by incorporating the newly distilled information from the node feature matrix $\mathbf{H}^{l-1}\in \mathbb{R}^{N\times d_H}$ at the ($l-1$)-th layer. Each of $N$ nodes has $d_H$ dimensions of features. $\mathcal{F}_\text{M}$ is the transformation function by the Graph Neural Memory that will be explained concretely in Section~\ref{sec:GNM}.
The obtained compact graph representation $\mathbf{M}^l$ is further leveraged to measure the global saliency of each node for graph classification. In particular, we learn global saliency by modeling the compatibility between $\mathbf{M}^l$ and node features via a function $\mathcal{F}_\text{S}$ (explicated in Section~\ref{sec:global_regular}):
\begin{equation}
    \mathbf{s}^{l-1} = \mathcal{F}_\text{S}(\mathbf{M}^l, \mathbf{H}^{l-1}).
    \label{eqn:compat}
\end{equation}
Herein, $\mathbf{s}^{l-1} \in \mathbb{R}^{N}$ is a vector, which sums to 1, denoting the saliency distribution for all nodes in the $(l-1)$-th layer of the backbone network. It can be interpreted as the semantic similarities  between each node and the graph representation. The saliency $\mathbf{s}^{l-1}$ is then used to regularize the backbone network and thereby refines the node features in the $l$-th layer:
\begin{equation}
    \mathbf{H}^{l}= \mathcal{F}_\text{B}(\mathbf{H}^{l-1}, \mathbf{s}^{l-1}),
    \label{eqn:backbone}
\end{equation}
where $\mathcal{F}_\text{B}$ is the transformation function of the backbone network. The refined node features $\mathbf{H}^{l}$ are in turn used to distill the compact graph representation in the next layer $\mathbf{M}^{l+1}$ as shown in Equation~\ref{eqn:fm}. Thus, the compact graph representation and the node features are refined interdependently. 
The compact graph representation in the last layer of the Graph Neural Memory $\mathbf{M}^{L+1}$ ($L$ is the number of feature-refining layers in the backbone network) is finally utilized for graph classification.

We will first elaborate on the design of the Graph Neural Memory, and then describe how to perform global regularization for the backbone network using the compact graph representation obtained from the Graph Neural Memory. Finally, we show that our \emph{SAR-GNN} can be optimized in an end-to-end manner as a whole.

\subsection{Graph Neural Memory}
\label{sec:GNM}
The Graph Neural Memory is designed to distill a compact latent representation for the entire graph from node features in the backbone network. 
As shown in Figure~\ref{fig:architecture}, the Graph Neural Memory learns a vectorial latent embedding $\mathbf{M} \in \mathbb{R}^{d_M}$ as the compact graph representation and refines it iteratively using cross-attention operation. Specifically, the initial latent embedding in the $0$-th layer (denoted $\mathbf{M}^0$) is parameterized as a learnable vector and initialized randomly. In each layer of the Graph Neural Memory, the latent embedding is refined by attending to the node features of the previous layer in the backbone to perform feature distillation with cross-attention operation. Formally, the transformation function of the Graph Neural Memory $\mathcal{F}_\text{M}$ in Equation~\ref{eqn:fm} is modeled as:
\begin{equation}
\label{eqn:memory}
\begin{split}
    & \mathbf{q}_m = \mathbf{M}^{l-1} \mathbf{W}^q, \ \mathbf{K} = \mathbf{H}^{l-1} \mathbf{W}^K, \ \mathbf{V} = \mathbf{H}^{l-1} \mathbf{W}^V, \\
    &\mathbf{M}^l = \mathcal{F}_\text{MLP}(\mathbf{M}'), \quad \mathbf{M}' = \text{softmax}(\frac{\mathbf{q}_m \mathbf{K}^\top}{\sqrt{d}}) \mathbf{V},
\end{split}
\end{equation}
where the latent embedding $\mathbf{M}^{l-1}$ in the $l$-1-th layer of the Graph Neural Memory serves as the query while the node feature matrix $\mathbf{H}^{l-1}$ in the ($l$-1)-th layer of the backbone network is used as both the key and value in cross-attention operation. $\mathbf{W}^q \in \mathbb{R}^{d_M \times d}$, $\mathbf{W}^K \in \mathbb{R}^{d_H \times d}$, $\mathbf{W}^V \in \mathbb{R}^{d_H \times d}$ are learnable parameter matrices for linear transformations. $\mathcal{F}_\text{MLP}$ is a Multilayer Perceptron (MLP) module consisting of two fully connected layers (with transformation matrices $\mathbf{W}^1_\text{MLP} \in \mathbb{R}^{d_M \times d}$ and $\mathbf{W}^2_\text{MLP} \in \mathbb{R}^{d \times d}$) with ReLU function. In our implementation, the transformation $\mathcal{F}_\text{M}$ described in Equation~\ref{eqn:memory} is typically performed $k$ iterations (tuned as a hyper-parameter) in a single layer of the Graph Neural Memory. Note that residual connections are applied between adjacent layers of the Graph Neural Memory.

The learned latent embedding $\mathbf{M}^l$ is in turn used to regularize the backbone network for refining the node feature matrix $\mathbf{H}^{l}$ (Section~\ref{sec:global_regular}). Thus the Graph Neural Memory is able to distill graph representations from newly refined node features in each layer instead of constant node information. \MR{Similar way of information distillation via cross attention has been previously adopted in Transformer-based models like BERT~\cite{devlin2018bert} and ViT~\cite{yuan2021tokens} that learn a `CLS' token for classification, DETR for object detection~\cite{DETR} and Perceiver for multi-modal feature learning~\cite{perceiver}.} Note that the initial latent embedding $\mathbf{M}^0$ is modeled as learnable parameters to learn an appropriate initial feature point that is compatible with the latent feature space in the backbone.
 
\subsection{Saliency-Aware Regularization of the Backbone}
\label{sec:global_regular}
The distilled compact graph representation from the Graph Neural Memory is used to perform global regularization on the backbone network, which enables the backbone network to perceive the global node saliency w.r.t. graph classification when learning node features. Specifically, we first utilize the compact graph representation to measure the global saliency for each node, and then regularize the backbone network using the node saliency distribution.

\noindent\textbf{Measuring the global node saliency.}
The node saliency w.r.t. graph classification is measured by modeling the compatibility between the compact graph representation and features of each node in the graph. A straightforward way is to calculate the similarity between them by dot product in a projected latent space. The obtained similarity scores for all nodes are then normalized. Thus, the compatibility function $\mathcal{F}_\text{S}$ in Equation~\ref{eqn:compat} is formulated as:
\begin{equation}
\begin{split}
    & \mathbf{q}_s = \mathbf{M}^{l} \mathbf{W}^q_s, \ \mathbf{K}_s = \mathbf{H}^{l-1} \mathbf{W}^K_s,\\
    &\mathbf{s}^{l-1} = \text{softmax}(\frac{\mathbf{q}_s \mathbf{K}_s^\top}{\sqrt{d_s}}), 
\end{split}
\label{eqn:saliency}
\end{equation}
where $\mathbf{W}^q_s \in \mathbb{R}^{d_M \times d}$ and $\mathbf{W}^K_s \in \mathbb{R}^{d_H \times d}$ are learnable parameter matrices to project the compact graph representation $\mathbf{M}^{l}$ and the node feature matrix $\mathbf{H}^{l-1}$ into the same latent space. Note that the scaling factor $\frac{1}{\sqrt{d}}$ is used to avoid the explosive growth of dot product between two vectors.

\noindent\textbf{Regularizing the backbone network with node saliency.}
Typical graph neural networks refine features of a node by performing weighted aggregation 
over (transformed) features of its neighboring nodes, where the weights are either derived based on the node degrees like GCN or basic uniform distributions such as GraphSAGE or GIN. These methods solely model the local dependencies between neighboring nodes and fail to incorporate the global node saliency w.r.t. graph classification. Intuitively, the nodes with higher saliency score should receive more attention than those with lower saliency score during information propagation between adjacent nodes. Hence, we regularize the backbone network with the learned node saliency distribution. 


We formulate the aggregating weights for the neighborhood of the $i$-th node (including itself) in the $l$-th layer as local weight distribution $\mathbf{a}_i^l \in \mathbb{R}^{|\mathcal{N}_i|+1}$, where $\mathcal{N}_i$ denotes the set of neighboring nodes of the $i$-th node. The learned global saliency $\mathbf{s}^l$ is viewed as the global weight distribution. Then we regularize the feature aggregation for each node in the backbone network by fusing the local and global weights together. We discuss two fusion mechanisms:
\begin{itemize}
\item \textbf{Weighted sum} over the local and global weights:
\begin{equation}
    \begin{split}
    &\mathbf{w}_i^l(j) = \text{softmax}\big(\mathbf{a}_i^l(j) + \beta \mathbf{s}^l(j)\big), \\
    &\forall j \in \{i\} \cup \mathcal{N}_i, 1\leq i \leq N,
    \end{split}
    \label{eqn:regular1}
\end{equation}
where $\mathbf{w}_i^l \in \mathbb{R}^{|\mathcal{N}_i|+1}$ is the regularized aggregating weights for the $i$-th node and its neighboring nodes, and $j$ is a node index. $\beta$ is a hyper-parameter tuned on a validation set. Softmax function is applied for normalization. 
\item \textbf{Scaling regularization} over the local weights by the global weights:
\begin{equation}
\begin{split}
    &\mathbf{w}_i^l(j) = \text{softmax}\big((1+\mathbf{s}^l(j))^\gamma \mathbf{a}_i^l(j)\big), \\
    &\forall j \in \{i\} \cup \mathcal{N}_i, 1\leq i \leq N,
    \end{split}
    \label{eqn:regular2}
\end{equation}
wherein, $\gamma>0$ is a hyper-parameter.
\end{itemize}

The proposed saliency-aware regularization mechanism is readily applicable to most graph neural networks.  In our implementation, the backbone network is instantiated with four classical types of graph neural networks: GCN, GraphSAGE, GIN and GAT. The transformation function $\mathcal{F}_B$ in Equation~\ref{eqn:backbone} is formulated correspondingly as follows.
\begin{itemize}
\item\textbf{GCN}, which takes into account the degree distribution when performing neighborhood aggregation:
\begin{equation}
    \begin{split}
    & \mathbf{a}^l = \tilde{\mathbf{D}}^{-\frac{1}{2}}\tilde{\mathbf{A}}\tilde{\mathbf{D}}^{-\frac{1}{2}},\\
    & \mathbf{H}^{l+1} = \sigma \left  (\mathbf{w}^l  \mathbf{H}^{l}\Theta^{l} \right).
    \end{split}
    \label{eqn:GCN}
\end{equation}
Herein, we consider $\mathbf{a}^l \in \mathbb{R}^{N\times N}$ as the local aggregating weight matrix in the renormalized form~\cite{GCN} derived from the adjacency matrix $\tilde{\mathbf{A}}$ and degree matrix $\tilde{\mathbf{D}}$. 
$\mathbf{w}^l \in \mathbb{R}^{N\times N}$ is the regularized aggregating weight matrix for all nodes, calculated by Equation~\ref{eqn:regular1} or~\ref{eqn:regular2}. $\Theta^{l}$ is the learnable parameter matrix for linear transformation in $l$-th layer and $\sigma$ is the activation function.
\item \textbf{GraphSAGE}. Taking the mean aggregator of GraphSAGE as an example, the local aggregating weight for all neighboring nodes is equal to 1. 
Thus, the features for $i$-th node $\mathbf{H}_i^{(l+1)}$ is refined as:
\begin{equation}
\resizebox{0.9\linewidth}{!}{$
    {\mathbf{H}_i}^{(l+1)}=\sigma  \Big(\Theta^{l} {\text{MEAN} }(\{\mathbf{w}_i^l(i)\mathbf{H}_i^{l}\}\cup \{ \mathbf{w}_i^l(j)\mathbf{H}_j^{l}, \forall j \in \mathcal{N}_i \}\Big).
    $}
\end{equation}
Here $\mathbf{w}_i^l$ is the regularized weights for the $i$-th node and its neighboring nodes by Equation~\ref{eqn:regular1} or~\ref{eqn:regular2}.
\item \textbf{GIN}, whose local aggregating weights are equal to 1 except for the center node:
    \begin{equation}
\begin{split}
    &\mathbf{a}^l(i) = 1+\epsilon^{l+1}, \quad \mathbf{a}^l(j) = 1, \forall j \in \mathcal{N}_i,\\
    &{\mathbf{H}_i}^{l+1}=\mathcal{F}_{\text{MLP}}^{l} \Big(\mathbf{w}_i^l(i)\mathbf{H}_i^{l}+\sum_{j \in \mathcal{N}_i}  \mathbf{w}_i^l(j)\mathbf{H}_j^{l} \Big),
    \end{split}
\end{equation}
where $\mathcal{F}_{\text{MLP}}^{l}$ denotes the transformation function for a Multilayer Perceptron at the $l$-th layer of backbone network, and $\mathbf{w}_i^l$ is the regularized weights for the $i$-th node and its neighboring nodes by Equation~\ref{eqn:regular1} or~\ref{eqn:regular2}.

\item \textbf{GAT}, which uses an attention mechanism to calculate the local weight distribution between each node and its neighboring nodes:
\begin{equation}
\begin{split}
&\mathbf{a}_i^l(j)=\frac{{\rm exp}({\rm LeakyReLU}(\mathbf{a}^\top [\mathbf{W}\mathbf{h}_{i}\parallel \mathbf{W}\mathbf{h}_{j} ]))}{\sum_{j\in \{i\} \cup \mathcal{N}(i)}{\rm exp}({\rm LeakyReLU}(\mathbf{a}^\top[\mathbf{W}\mathbf{h}_{i}\parallel \mathbf{W}\mathbf{h}_{j} ]))}, \\
&\mathbf{h}_{i}^{l+1}=\sigma(\sum_{j\in \{i\} \cup \mathcal{N}(i)}\mathbf{w}_i^l(j)  \mathbf{W}\mathbf{h}_j).
\end{split}
\end{equation}
where $\mathbf{h}_{i}$ are the feature vector for the $i$-th node and $\mathbf{W}$ is a shared transformation matrix. $\mathbf{a}$ is the parameters for a linear transformation to calculate a scalar matching score while $\parallel$ denotes the concatenation operation. $\mathbf{w}_i^l(j)$ is the regularized weights between the $i$-th node and the $j$-th node.
\end{itemize}

\subsection{End-to-End Parameter Learning}
Since the compact graph representation in the $1$-th layer of the Graph Neural Memory $\mathbf{M}^1$ is learned from the input node features $\mathbf{X}$, the Graph Neural Memory contains one more layer than the backbone network. Suppose the backbone network contains in total $L$ feature-refining layers, the learned compact graph representation in the last layer of the Memory $\mathbf{M}^{L+1}$ is utilized for predicting the graph label:
\begin{equation}
    P(\mathbf{y}|\mathbf{M}^{L+1}) = \text{softmax}(\mathcal{F}_{\text{MLP}}(\mathbf{M}^{L+1})),
\end{equation}
where $\mathbf{y} \in \mathbb{R}^K$ is predicted probabilities for all potential $K$ categories. $\mathcal{F}_{\text{MLP}}$ refers to the transformation function for a Multilayer Perceptron.

The proposed \emph{SAR-GNN} can be optimized in an end-to-end manner by a cross-entropy loss:
\begin{equation}
    \mathcal{L} = - \sum_{i=1}^{N_t} \log P(\mathbf{y}^i | \mathbf{X}^i).
\end{equation}

Here $\mathbf{X}^i$ and $\mathbf{y}^i$ refer to the $i$-th training sample with its graph label while $N_t$ is the training size. 
\section{Experiments}
To evaluate our proposed \emph{SAR-GNN}, we first conduct ablation study to investigate the effectiveness of each essential technique in our model, then we conduct extensive experiments to compare our model with other state-of-the-art methods for graph classification on seven challenging graph datasets across various types of graph data. Moreover, we also perform qualitative evaluation to validate the effectiveness of the learned node saliency and the learned graph representation.

\subsection{Experimental Setup}
\label{sec:setup}
\noindent\textbf{Datasets.} We conduct experiments on seven datasets across various types of graph data for evaluation. 1) \emph{MUTAG}~\cite{mutag} is a chemical dataset containing  two categories of compounds: mutagenic aromatic and heteroaromatic compounds. 2) \emph{ENZYMES}~\cite{enzymes} is a protein dataset comprising 600 protein tertiary structures, which are grouped into 6 top-level enzyme classes. 3) \emph{PROTEINS}~\cite{borgwardt2005protein} is also a protein dataset which contains two types of samples: enzymes or non-enzymes. 4-5) \emph{IMDB-MULTI}~\cite{yanardag2015deep} and \emph{IMDB-BINARY}~\cite{yanardag2015deep} both include ego-network graphs involving 1000 actors who play roles in movies appeared in IMDB. All graphs in \emph{IMDB-MULTI} are categories into three genre classes including Comedy, Romance and Sci-Fi while \emph{IMDB-BINARY} has two categories: Action and Romance. 6) \emph{TRIANGLES}~\cite{knyazev2019understanding} is a large synthetic dataset consisting of 45,000 graph samples, and the task is to predict the number of triangles in a graph (ranging from 1 to 10). Note that we perform evaluation on the split of the test set, which has the same number of categories of graphs as the training set. 7) \emph{Letter-high}~\cite{letterdata} is also a synthetic dataset containing 2,250 graph samples and each sample represents a distorted letter drawings.


\noindent\textbf{Implementation details.}
\label{sec:implementation}
Following typical evaluation protocol for graph classification~\cite{Errica2020A}, we perform 10-fold cross validation for all datasets except for \emph{TRIANGLES} dataset which has the official data division for training and test. For quantitative evaluation, we measure the accuracy of graph classification as the evaluation metric. 
Adam~\cite{adam} is employed for gradient descent optimization.

\subsection{Ablation Study}

To investigate the effectiveness of each proposed component, we first perform ablation study with six variants of our \emph{SAR-GNN} for all four types of backbones. 
\begin{itemize}
\item \textbf{Base model}, which is exactly the backbone network, e.g., base model is GCN when it is used as the backbone. Following Errica et al.~\cite{Errica2020A}, we apply sum pooling to obtain graph representations for all \emph{Base} models. 
\item \textbf{GNM-GNN}, which employs the Graph Neural Memory (GNM) to learn the compact graph representation for graph classification. Nevertheless, the backbone is not regularized with the node saliency. 
\item \textbf{SAR-Pooling}, which performs sum pooling over all node features in the last layer of the backbone for graph classification, instead of using the compact graph representation. Note that the backbone of this variant is regularized with node saliency. In particular, two proposed mechanisms for weight fusion (Equations~\ref{eqn:regular1} and \ref{eqn:regular2}), namely `weighted sum' and `scaling regularization', are evaluated separately. We denote them as \textbf{SAR-Pooling-W} and \textbf{SAR-Pooling-S} respectively. 
\item \textbf{SAR-GNN}, which is the intact version of our model. Both mechanisms for weight fusion are evaluated, denoted as \textbf{SAR-GNN-W} and \textbf{SAR-GNN-S} respectively. 
\end{itemize}
Table~\ref{tab:ablation} shows the experimental results of these variants on \emph{MUTAG} and \emph{Letter-high} datasets, from which we conduct following investigation. 

\begin{table}
\caption{Classification accuracy ($\%$) of six variants of our model on \emph{MUTAG} and \emph{Letter-high} for ablation study. The backbone is instantiated with GCN, GIN, GraphSAGE and GAT, respectively. The standard deviation on 10-fold cross validation is also provided.}
\vspace{-5pt}
\label{tab:ablation}
\begin{center}
\begin{small}
\renewcommand\arraystretch{1.4}
\resizebox{1.\linewidth}{!}{
\begin{tabular}{l|cc|cc|cc|cc}
\toprule
\multirow{2}{*}{\diagbox[]{Variants}{Backbone}} & \multicolumn{2}{c|}{GCN} & \multicolumn{2}{c|}{GIN} & \multicolumn{2}{c|}{GraphSAGE} & \multicolumn{2}{c}{GAT} \\
 & MUTAG & Letter-high & MUTAG & Letter-high & MUTAG & Letter-high & MUTAG & Letter-high\\
\midrule
Base model   & 71.6 $\pm $10.9 & 61.1$\pm $2.6& 81.4$\pm$6.6 & 73.3$\pm$2.5& 75.8$\pm$7.8 & 71.0$\pm$2.7 & 76.1 $\pm $4.2 & 82.1$\pm $1.8\\ 
GNM-GNN &76.7$\pm$7.0 & 73.2$\pm$1.9 & 84.7$\pm$5.6  & 79.5$\pm$2.2& 78.4$\pm$7.3 & 75.9$\pm$2.3 & 77.5 $\pm $4.3 & 82.6$\pm $1.3\\
SAR-Pooling-W & 76.2$\pm$6.0 &70.7$\pm$1.8 & 83.8$\pm$4.0 & 75.3$\pm$1.8& 82.8$\pm$6.8 & 76.8$\pm$2.5 & 79.9 $\pm $2.7 & 83.4$\pm $4.6\\
SAR-Pooling-S &75.9$\pm$10.1& 70.5$\pm$2.9& 84.0$\pm$7.3 & 74.1$\pm$3.4 & 81.9$\pm$10.3 & 75.8$\pm$2.9 & 78.3 $\pm $3.2 & 84.4$\pm $4.1\\
SAR-GNN-W & 80.0$\pm$5.1 & 77.6$\pm$2.4& 85.1$\pm$5.0& 82.3$\pm$1.8& 84.6$\pm$4.2 & 80.8$\pm$2.6 & 81.4 $\pm $2.8 & 85.1$\pm $2.6\\
SAR-GNN-S & 82.2$\pm$5.5 & 76.8$\pm$1.7& 85.3$\pm$5.5 & 83.6$\pm$1.7& 87.4$\pm$3.1 & 80.5$\pm$2.6 & 80.2 $\pm $3.4 & 85.0$\pm $5.2\\
\bottomrule
\end{tabular}
}
\end{small}
\end{center}

\end{table}

\smallskip\noindent\textbf{Effect of regularization with learned node saliency.}
The large performance gap between \emph{Base} model and \emph{SAR-Pooling-W} as well as \emph{SAR-Pooling-S} for all four backbones on both datasets reveals the effect of the global regularization with learned node saliency. Besides, the comparisons between \emph{GNM-GNN} and \emph{SAR-GNN} also demonstrate the advantages of such saliency-aware regularization.

\smallskip\noindent\textbf{Effect of the Graph Neural Memory.}
Compared with \emph{Base} model, \emph{GNM-GNN} improves the performance by a large margin for all types of backbones on both datasets. Additionally, \emph{SAR-GNN} consistently outperforms \emph{SAR-Pooling} using the same fusion mechanism in all cases (with different backbones and datasets). These results validate the superiority of the Graph Neural Memory over the pooling method in learning graph representation.

\smallskip\noindent\textbf{Comparison between `Weighted sum' and `Scaling regularization' for weight fusion.}
We further compare two different mechanisms for weight fusion in Equations~\ref{eqn:regular1} and~\ref{eqn:regular2}. Comparing the performance between \emph{SAR-GNN-W} and \emph{SAR-GNN-S}, or between \emph{SAR-Pooling-W} and \emph{SAR-Pooling-S} in all cases, we observe that there is no clear winner between these two fusion mechanisms. In most cases, there is no big performance gap between them.


\smallskip\noindent\textbf{\MR{Investigation into the optimization policy.}} \MR{our \emph{SAR-GNN} consists of two core modules: the backbone network for learning node features and Graph Neural Memory for distilling the compact graph representation. Two modules work interdependently to refine each other and the whole model can be optimized in an end-to-end manner. We investigate two optional optimization policies: 1) joint training of two modules and 2) alternating training of two modules.}

\begin{itemize}
    \item \textbf{\MR{Joint training,}} \MR{which optimizes two modules jointly by gradient descent method. Adam is employed in our implementation.}
    \item \textbf{\MR{Alternating training,}} \MR{in which two modules are optimized in an alternating manner, i.e., the parameters of one module are frozen when optimizing the other one. Such policy is essentially analogous to the block coordinate descent method when considering the two modules as two blocks of parameters to be optimized. The only difference is that we still employ gradient descent method to optimize each block of parameters instead of calculating the local optimal solution per block directly. Similar to the block coordinate descent method, we optimize two modules by conducting such alternating training iteratively to reach the learning convergence.}
\end{itemize}

\begin{figure}
\centering
\subfigure[MUTAG] {\includegraphics[width=0.4\textwidth]{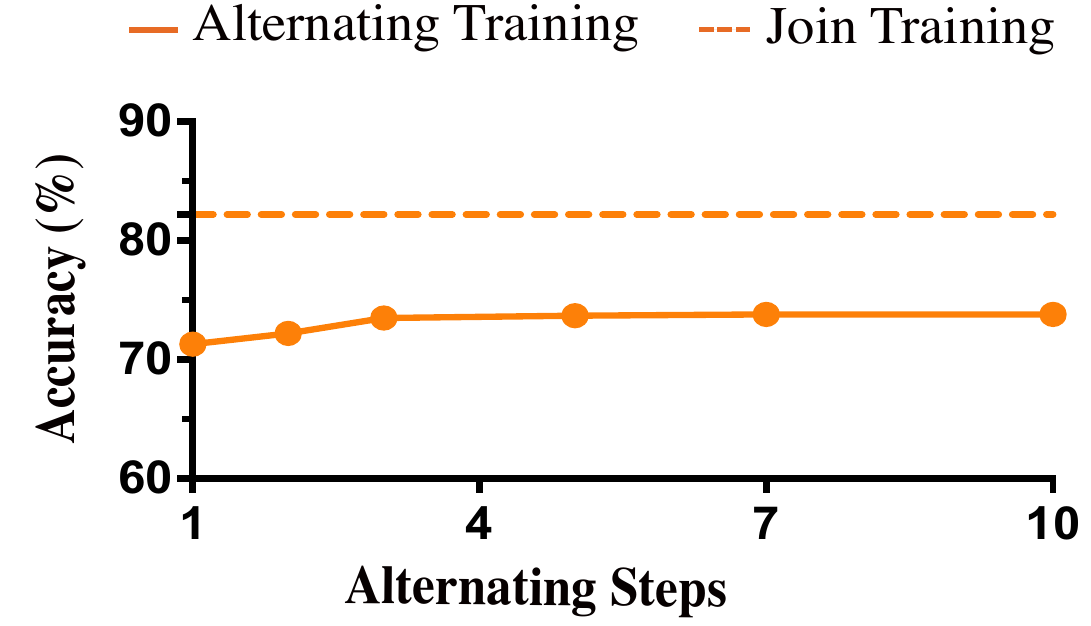}}
\subfigure[Letter-high] {\includegraphics[width=0.4\textwidth]{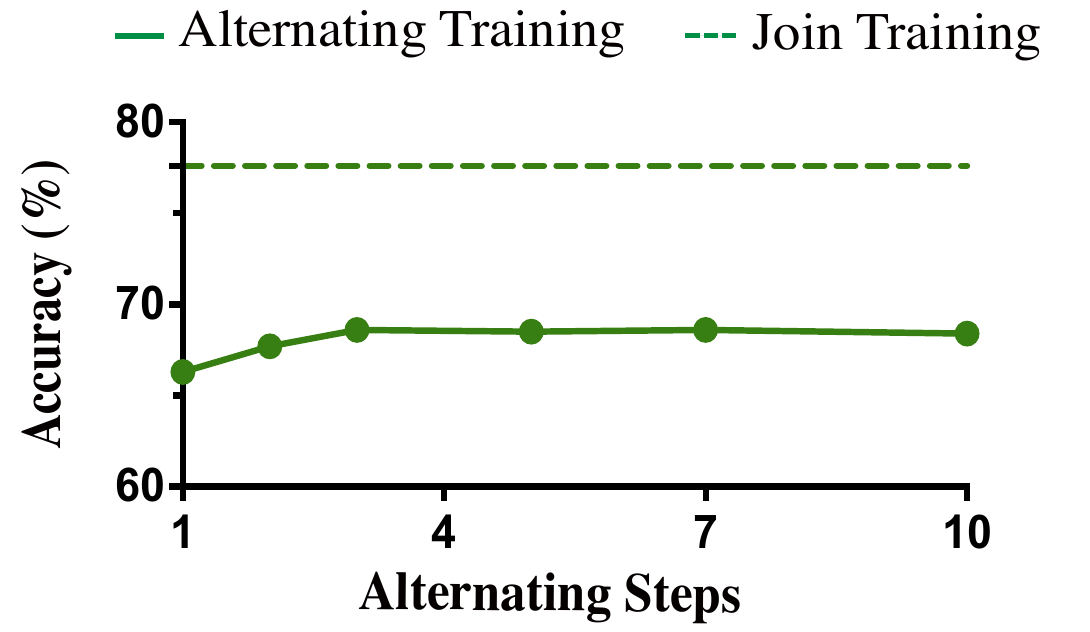}}
\caption{\MR{The performance of our \emph{SAR-GNN} adopting two optimization policies, namely the joint training policy and the alternating training policy, on MUTAG (a) and Letter-high datasets (b), respectively.}}
\label{fig:alter}
\end{figure}

\MR{Figure~\ref{fig:alter} presents the experimental comparison between two optimization policies. The alternating training policy reaches the optimization convergence after several iterations of alternating optimization. 
We observe that the joint training policy achieves better performance for graph classification than the alternating training policy. Similar to the coordinate descent method, the alternating training policy can theoretically achieve globally optimum performance if it can obtain the local optimum w.r.t. the optimized module in each iteration of per-module optimization for convex optimization problem. However, in our case both alternating training and joint training policies employ gradient descent for non-convex optimization of deep neural networks. Thus, the alternating training policy can hardly reach the optimum w.r.t. the optimized module in each alternating optimization step in practice. On the other hand, the joint training policy performs gradient back-propagation w.r.t. all parameters of the model whilst alternating training only propagates gradients w.r.t. the parameters of the optimized module of the model for optimization. Thus, we surmise that joint training policy is able to optimize the model along more optimized directions than alternating training, thereby yielding better performance.}

\subsection{Comparison with Other Methods}

Next we compare our model with the state-of-the-art methods for graph classification. Apart from four backbone models (GCN, GIN, GraphSAGE and GAT), we also compare our model with \MR{12} state-of-the-art methods for graph classification, 
including DGCNN~\cite{DGCNN}, ECC~\cite{ECC}, ChebyGIN~\cite{knyazev2019understanding}, SAGPool~\cite{lee2019self}, Graphormer~\cite{ying2021transformers}, ESAN~\cite{ESAN} (termed `DSS-GNN' in the paper) based on the overall best policy `EGO+', GSN~\cite{GSN}, Graph U-Nets~\cite{gao2019graph}, DGM~\cite{DGM}, UGformer~\cite{UGF}, \MR{SLIM~\cite{SLIM} and ADSF-RWR~\cite{ADSF}}. \MR{ADSF-RWR is initially designed for node classification and we adapt it for graph classification by applying sum pooling over all node embeddings to obtain the graph representation. We re-train and re-evaluate all methods under the same experimental setting, using the same data split and evaluation metrics, to have a fair comparison. Note that we compare with ChebyGIN in both unsupervised (denoted as ChebyGIN-unsup) and supervised training settings for learning node saliencies to have a comprehensive comparison.} In this set of experiments, we only report the performance of our model by selecting the better one between two fusion mechanisms based on the validation set.

\begin{table}
\caption{Classification accuracy ($\%$) of different methods for graph classification on seven datasets. The standard deviations on 10-fold cross validation (different initializations for \emph{TRIANGLES}) are provided. 
`SAR-GCN' refers to our method using GCN as the backbone. The performance gains of our method over each of four \emph{base} models are highlighted in \textit{italic} in parentheses.}
\label{tab:all-comparison}
\begin{center}
\renewcommand\arraystretch{1.2} 
\resizebox{1.0\linewidth}{!}{
\begin{tabular}{l|lllllll}
\toprule
Method   & MUTAG &ENZYMES & IMDB-MULTI &TRIANGLES & Letter-high & PROTEINS & IMDB-BINARY \\
\midrule
DGCNN~\cite{DGCNN} & 83.9$\pm$5.8  & 38.6$\pm$4.8 & 46.4$\pm$3.5& 77.3$\pm$4.2 & 44.7$\pm$2.9 & 72.5$\pm$4.1 & 66.9$\pm$3.9 \\
ECC~\cite{ECC}  & 81.4$\pm$7.6 & 29.5$\pm$8.2 &43.5$\pm$3.1 & 80.4$\pm$0.5 & 80.8$\pm$1.5 & 72.3$\pm$4.0 & 66.2$\pm$2.1 \\
ChebyGIN-unsup~\cite{knyazev2019understanding} & 83.4$\pm$6.0 & 26.3$\pm$2.7 &39.0$\pm$1.5 & 67.0$\pm$3.0 & 27.0$\pm$2.2 & 71.2$\pm$3.4 & 50.0$\pm$0.5\\
DSS-GNN (EGO+)~\cite{ESAN} &  84.1$\pm$5.9 & 62.5$\pm$5.7 & 50.1$\pm$2.4 & \textbf{93.3}$\pm$1.1& 83.6$\pm$2.0 & 69.2$\pm$5.6 & 71.5$\pm$3.6\\
GSN~\cite{GSN} &  84.4$\pm$5.5 & 63.2$\pm$6.8 & 51.1$\pm$2.7 & 76.1$\pm$5.2&40.3$\pm$2.4 & 72.9$\pm$3.7 & 72.5$\pm$4.1\\
Graphormer~\cite{ying2021transformers} & 83.1$\pm$6.7 & 58.7$\pm$5.2 &48.9$\pm$2.7 & 56.1$\pm$2.1 & 76.3$\pm$2.4 & 76.3$\pm$3.4 & 72.9$\pm$3.3 \\
SAGPOOL~\cite{lee2019self}  & 72.6$\pm$5.7 & 31.9$\pm$5.7 &46.0$\pm$3.8 & 63.2$\pm$10.3 & 72.2$\pm$3.3 & 72.6$\pm$3.3 & 69.2$\pm$3.5 \\
Graph U-Nets~\cite{gao2019graph} & 76.0$\pm$7.2 & 34.4$\pm$3.0 &47.6$\pm$3.0 & 54.0$\pm$1.0 & 42.5$\pm$1.3 & 74.2$\pm$5.2 & 68.3$\pm$3.3 \\
DGM~\cite{DGM} & 86.2$\pm$6.7 & 40.3$\pm$2.2 &49.2$\pm$2.5 & 53.4$\pm$2.3 & 56.4$\pm$3.4 & 68.8$\pm$6.9 & 73.1$\pm$3.8 \\
UGformer~\cite{UGF} & 78.2$\pm$ 5.9&  68.9$\pm$4.7 & 49.6$\pm$ 2.1& 81.0$\pm$4.6 & 80.6$\pm$2.7 & 70.1$\pm$3.5 & 70.1$\pm$2.3 \\
\MR{SLIM~\cite{SLIM}} & \MR{78.8$\pm$ 4.8}&  \MR{56.1$\pm$3.4} & \MR{47.6$\pm$ 2.1}& \MR{68.9$\pm$3.1} & \MR{71.3$\pm$1.8} & \MR{74.1$\pm$4.7} & \MR{70.8$\pm$5.7} \\

\MR{ADSF-RWR~\cite{ADSF}} & \MR{76.4$\pm$ 6.2} &  \MR{66.0$\pm$4.3 }&\MR{45.2$\pm$ 4.9}& \MR{62.4$\pm$5.2}& \MR{80.1$\pm$2.4}& \MR{73.9$\pm$6.8} & \MR{69.4$\pm$2.7} \\
\midrule
GCN & 71.6$\pm$10.9 & 68.1$ \pm$4.0 &48.2$\pm$3.8 & 46.0$\pm$1.0 & 61.1$\pm$2.6 & 73.3$\pm$3.1 & 68.4$\pm$6.7 \\
GIN & 81.4$\pm$6.6 & 62.3$\pm$5.0 &48.5$\pm$3.3& 51.6$\pm$2.7 & 73.3$\pm$2.5. & 73.5$\pm$3.4 & 70.2$\pm$2.8\\
GraphSAGE & 75.8$\pm$7.8  & 62.1$ \pm$6.2 &47.6$\pm$3.5 & 52.5$\pm$0.4 & 71.0$\pm$2.7 & 73.0$\pm$4.6 & 70.7$\pm$4.4\\
GAT & 76.1$\pm$4.2  &65.4 $ \pm$ 5.2&46.0$\pm$3.2 & 65.4$\pm$3.4 & 82.1$\pm$1.8 & 73.5$\pm$2.6 & 70.0$\pm$5.1\\
\midrule
SAR-GCN (Ours) & 82.2$\pm$5.5 (\textit{+10.6}) & \textbf{69.3}$\pm$5.0  (\textit{+1.2}) & 49.1$\pm$3.1 (\textit{+0.9}) & 81.7$\pm$0.9 (\textit{+35.7}) & 77.6$\pm$2.4  (\textit{+16.5}) &  \textbf{76.8}$\pm$3.4 (\textit{+3.5})& 69.2$\pm$4.8 (\textit{+0.8})\\
SAR-GIN (Ours) & 85.3$\pm$5.5 (\textit{+3.9})  & 65.7$\pm$3.5 (\textit{+3.4}) & \textbf{51.6}$\pm$3.7 (\textit{+3.1})& 78.2$\pm$4.2 (\textit{+26.6}) &\textbf{84.8}$\pm$1.3 (\textit{+11.5})  &  75.3$\pm$6.4 (\textit{+1.8}) & \textbf{73.7}$\pm$3.9 (\textit{+3.5})\\
SAR-GraphSAGE (Ours) & \textbf{87.4}$\pm$3.1 (\textit{+11.6})& 68.7$\pm$4.0 (\textit{+6.6})&48.7$\pm$3.2 (\textit{+1.1}) & 78.9$\pm$1.2 (\textit{+26.4}) & 80.5$\pm$2.6 (\textit{+9.5})  &  76.0$\pm$4.7  (\textit{+3.0}) & 70.9$\pm$4.1 (\textit{+0.2}) \\
SAR-GAT (Ours) & 81.4 $\pm $2.8 (\textit{+5.3})& $69.1\pm$4.7 (\textit{+3.7})&47.7$\pm$2.6 (\textit{+1.7}) & 87.4$\pm$4.6 (\textit{+22.0}) & 85.1$\pm$2.6 (\textit{+3.0})  & 75.8 $\pm$1.9  (\textit{+2.3}) & 72.5$\pm$2.8 (\textit{+2.5}) \\
\bottomrule
\end{tabular}
}
\end{center}
\end{table}

\begin{figure}[!t]
\centering
\includegraphics[width=0.9\linewidth]{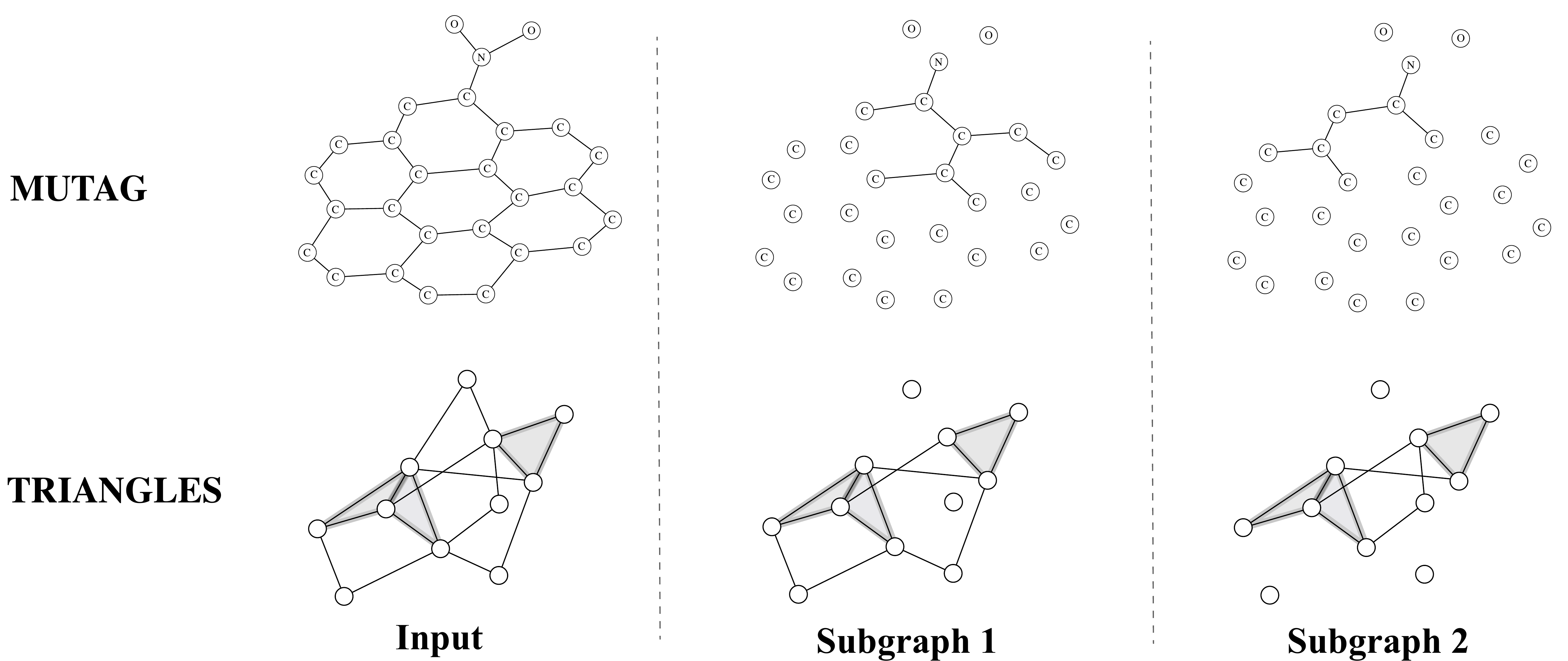}
\caption{\MR{Visualization of two randomly selected subgraphs generated by DSS-GNN (EGO+) on samples from MUTAG and TRIANGLES datasets, respectively. The triangles are indicated in shaded regions for the TRIANGLES sample. While DSS-GNN can preserve the triangle structure in the generated subgraphs, its generated subgraphs on the MUTAG sample fail to preserve intact fused rings whose number is closely correlated to the recognition of mutagenic aromatic.}}
\label{fig:dss-subgraph}
\end{figure}

Table~\ref{tab:all-comparison} presents the experimental results of different methods on seven datasets. We make the following observations. First, our method achieves substantial performance gains compared to the corresponding base models used for backbone on all datasets except \emph{IMDB-BINARY}, which validates the effectiveness and adaptability of our method across different types of backbones. The performance improvement of our method on \emph{IMDB-BINARY} is minor presumably due to the ceiling effect, considering the fact that the performance of multiple models on this dataset is close to each other. Second, our method based on one of four backbones achieves the best performance on all seven datasets except TRIANGLES, which manifests the robustness of our method. In contrast, other methods, such as GSN, Graphormer and ECC, show unstable performance across different datasets. DSS-GNN performs substantially better than other methods on \emph{TRIANGLES}, presumably because DSS-GNN can recognize the triangle structure as distinguishable subgraphs. \MR{To validate it, we visualize two randomly selected subgraphs learned by DSS-GNN in Figure~\ref{fig:dss-subgraph}, which shows that DSS-GNN can indeed generate subgraphs that preserve the triangles on the samples from TRIANGLES dataset. However, its generated subgraphs on the MUTAG sample fail to preserve intact fused rings whose number is closely correlated to the recognition of mutagenic aromatic~\cite{mutag}, which accounts for its relatively inferior performance to our model on MUTAG.} The third observation we make is that our method consistently performs well when built upon different backbones and the performance differences between them are minor. Interestingly, the performance differences among four base models are substantially reduced when they are integrated into our method.

\smallskip\noindent\textbf{\MR{Comparison with the implicit modeling of global saliency by attention mechanism.}} \MR{SAGPool and ChebyGIN use attention mechanism to model global saliency implicitly by learning a parameterized attention function $\mathcal{F}_\text{att}$ taking the node features $\mathbf{H}$ as input: $\mathbf{s}_\text{att} = \mathcal{F}_\text{att}(\mathbf{H}).$ In contrast, our method models the global saliency explicitly by measuring the compatibility between the compact graph representation $\mathbf{M}$ and node features $\mathbf{H}$: $\mathbf{s}_\text{our} = \mathcal{F}_\text{S}(\mathbf{M}, \mathbf{H})$, as shown in Equation~\ref{eqn:compat} and~\ref{eqn:saliency}. The iterative interdependent refining mechanism enables our model to learn more effective node features and graph representation, which in turn lead to more accurate global saliency. The large performance superiority of our model over SAGPool and ChebyGIN in Table~\ref{tab:all-comparison} in the paper demonstrates the advantage of our model. Besides, the qualitative comparison in Figure~\ref{fig:saliency_triangle} and Figure~\ref{fig:saliency_mutag} also reveal such advantage of our model over SAGPool and ChebyGIN.} 

\smallskip\noindent\textbf{\MR{Comparison with `ChebyGIN' in the supervised setting for learning global node saliency.}} \MR{Our model, as well as other methods involved in comparison, is optimized for graph classification only using graph category labels but with no annotations of the global saliency. In this sense, the global saliency is learned in an unsupervised setting. To further investigate the performance of the our model in the supervised setting, we conduct experiments on TRIANGLES dataset by leveraging both the annotations of the global saliency and graph categories for supervision. We compare our method and the supervised version of ChebyGIN, which is presented in Table~\ref{tab:supervised}. The results show that all four instantiations of our method with different backbones consistently outperform ChebyGIN in the supervised setting, which reveals the effectiveness of our method in such setting. Besides, comparing the performance between unsupervised and supervised settings, both our method and ChebyGIN achieve substantial improvements. Such results indicate that the supervision on the global saliency yields more effective learning of global saliency and thereby more precise graph classification, which also implies the essential benefit of learning global node saliencies to the task of graph classification.}

\begin{table}[tb]
\caption{\MR{Classification accuracy ($\%$) of our methods instantiated with different backbones as well as ChebyGIN in both unsupervised and supervised learning settings on TRIANGLES dataset.}} 
\label{tab:supervised}
	\begin{center}
\renewcommand\arraystretch{1.4}
\resizebox{0.8\linewidth}{!}{
					\begin{tabular}{l|c|cccc}
						\toprule
      \multirow{2}{*}{Learning settings}&\multirow{2}{*}{ChebyGIN}& \multicolumn{4}{|c}{Our method} \\
						 &   & SAR-GCN & SAR-GIN  &SAR-GraphSAGE & SAR-GAT \\
						\midrule
                        Unsupervised & 67.0$\pm$3.0  & 81.7$\pm$0.9 & 78.2$\pm$4.2 & 78.9$\pm$1.2 & 87.4$\pm$4.6\\
					Supervised&	\MR{88.0$\pm$1.0} &\MR{91.2$\pm$1.7} & \MR{89.8$\pm$2.1} & \MR{89.3$\pm$2.0} &\MR{91.8$\pm$1.5} \\
						\bottomrule
					\end{tabular}
				}
\end{center}
\end{table}

\subsection{Qualitative Evaluation}
In this set of experiments, we perform qualitative evaluation on the learned node saliency and the graph representation by our model, respectively.

\begin{figure}[!t]
\centering
 \includegraphics[width=1\linewidth]{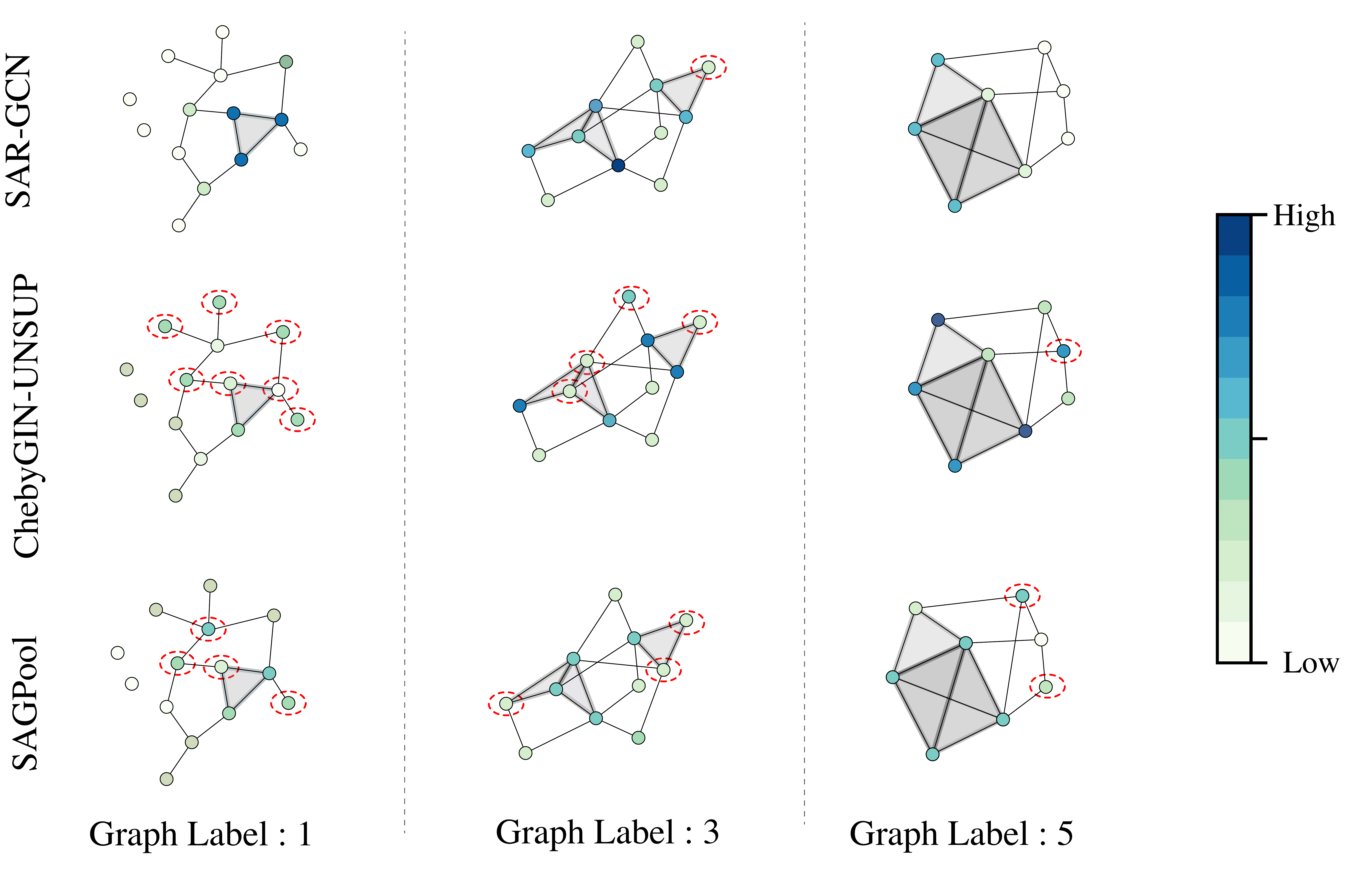}
\caption{Visualization of learned node saliency by our SAR-GCN, ChebyGIN-unsup and SAGPool on three graphs randomly selected from TRIANGLES test data. The triangles are indicated in shaded regions and the number of triangles is given as the graph label. Our SAR-GCN tends to assign higher saliency scores to those nodes associated with more triangles, and thus shows higher accuracy than ChebyGIN-unsup and SAGPool in capturing the discriminative nodes associated with more triangles. The nodes that are assigned distinctly inaccurate saliency weights are marked with red dotted circles.}
\label{fig:saliency_triangle}
\end{figure}

\begin{figure}[!t]
\centering
\includegraphics[width=1\linewidth]{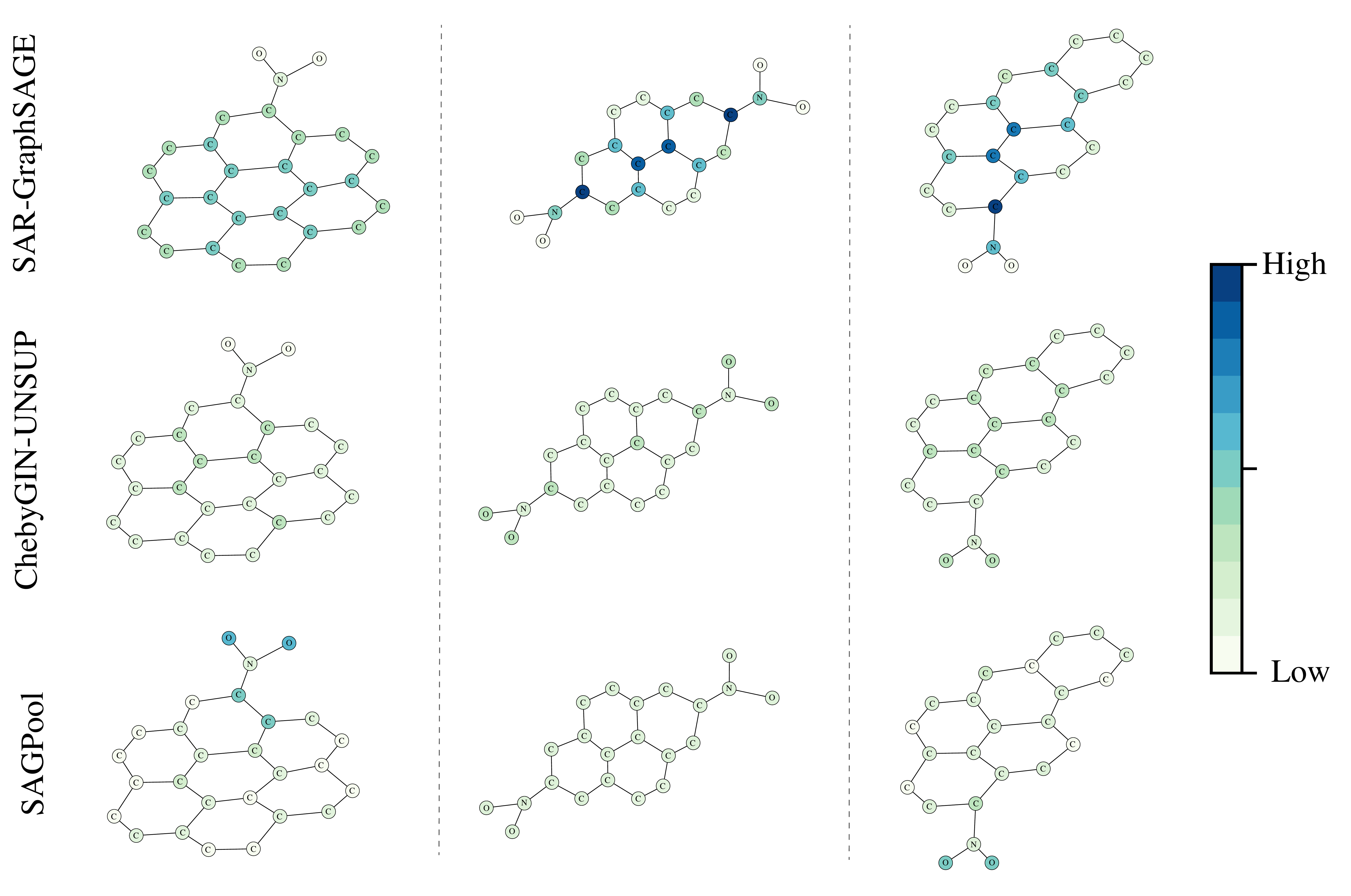}
\caption{Visualization of learned node saliency by our SAR-GraphSAGE, ChebyGIN-unsup and SAGPool on three graphs, labeled as `mutagenic aromatic', randomly selected from MUTAG. Our SAR-GraphSAGE can capture the key nodes located at the center of fused rings which are discriminative for `mutagenic aromatic', whilst ChebyGIN-unsup and SAGPool either assign roughly uniform attention to all nodes or fail to capture the key nodes.}
\label{fig:saliency_mutag}
\end{figure}

\smallskip\noindent\textbf{Evaluation of the learned node saliency.} To validate whether the learned node saliency by our method can accurately reflect the relevance of each node to the task of graph classification, we visualize the learned node saliency on \emph{TRIANGLES} and \emph{MUTAG} datasets in Figures~\ref{fig:saliency_triangle} and~\ref{fig:saliency_mutag}, respectively.
We also visualize the learned global attention scores by ChebyGIN-unsup and SAGPool which employs an attention mechanism to model the global saliency in an implicit manner. 

As shown in Figure~\ref{fig:saliency_triangle}, our method tends to assign higher saliency scores to those nodes associated with more triangles, which is reasonable since these nodes are more relevant to the graph category, namely the number of triangles in the graph. Although the learned attention scores by ChebyGIN-unsup and SAGPool also show a similar pattern, the results are less accurate than those of our method. 

Figure~\ref{fig:saliency_mutag} visualizes the learned saliency by our model as well as ChebyGIN-unsup and SAGPool on three positive samples labeled as `mutagenic aromatic', randomly selected from \emph{MUTAG}. Our model assigns higher saliency scores to the key nodes which locate at the center of fused rings, which is consistent with the chemical knowledge that `compounds with three or more fused rings are much more mutagenic, other factors being equal, than those with one or two'~\cite{mutag}. In contrast, ChebyGIN-unsup and SAGPool either learn a uniform distribution of attention weights for all nodes or fail to capture those key nodes. These visualizations clearly show the superiority of our method over ChebyGIN-unsup in learning the global node saliency for graph classification.

\begin{figure}[!t]
\centering
 \includegraphics[width=1\linewidth]{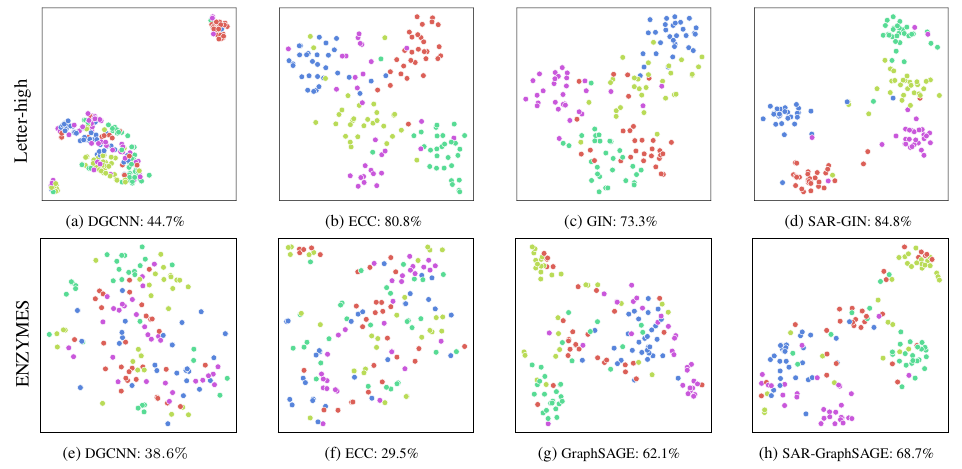}
\caption{t-SNE maps of test data from five randomly selected classes of \emph{Letter-high} and \emph{ENZYMES}, constructed based on the learned graph representations by four different methods. We show results of our model with different backbones (GIN and GraphSAGE) on two datasets to show the robustness of our model. 30 randomly selected samples are visualized for each class. The classification accuracy for each method on each dataset is also presented.}
\label{fig:tsne}
\end{figure}

\smallskip\noindent\textbf{Evaluation of the learned graph representation.} To gain further insight into the quality of the learned global graph representation by our method and other methods, we show t-SNE~\cite{tsne} maps of test data from \emph{Letter-high} and \emph{ENZYMES} datasets in Figure~\ref{fig:tsne}, constructed on the learned graph representations by four methods. We show results of our model with different backbones (GIN and GraphSAGE) on two datasets to show the robustness of our model. More effective graph representations typically lead to more separable clustering between classes in the t-SNE map. These maps reveal the consistent results with the quantitative evaluation among different methods in Section 4.3, which demonstrates 
that our method is able to learn more effective graph representation than other methods.

\subsection{\MR{Model Complexity}} 
\MR{We conduct extensive investigation into the model complexity of our model. In particular, we evaluate the model complexity in terms of model size and computational complexity, respectively.}

\begin{figure}[!t]
\centering
\subfigure[] {\includegraphics[width=0.5\textwidth]{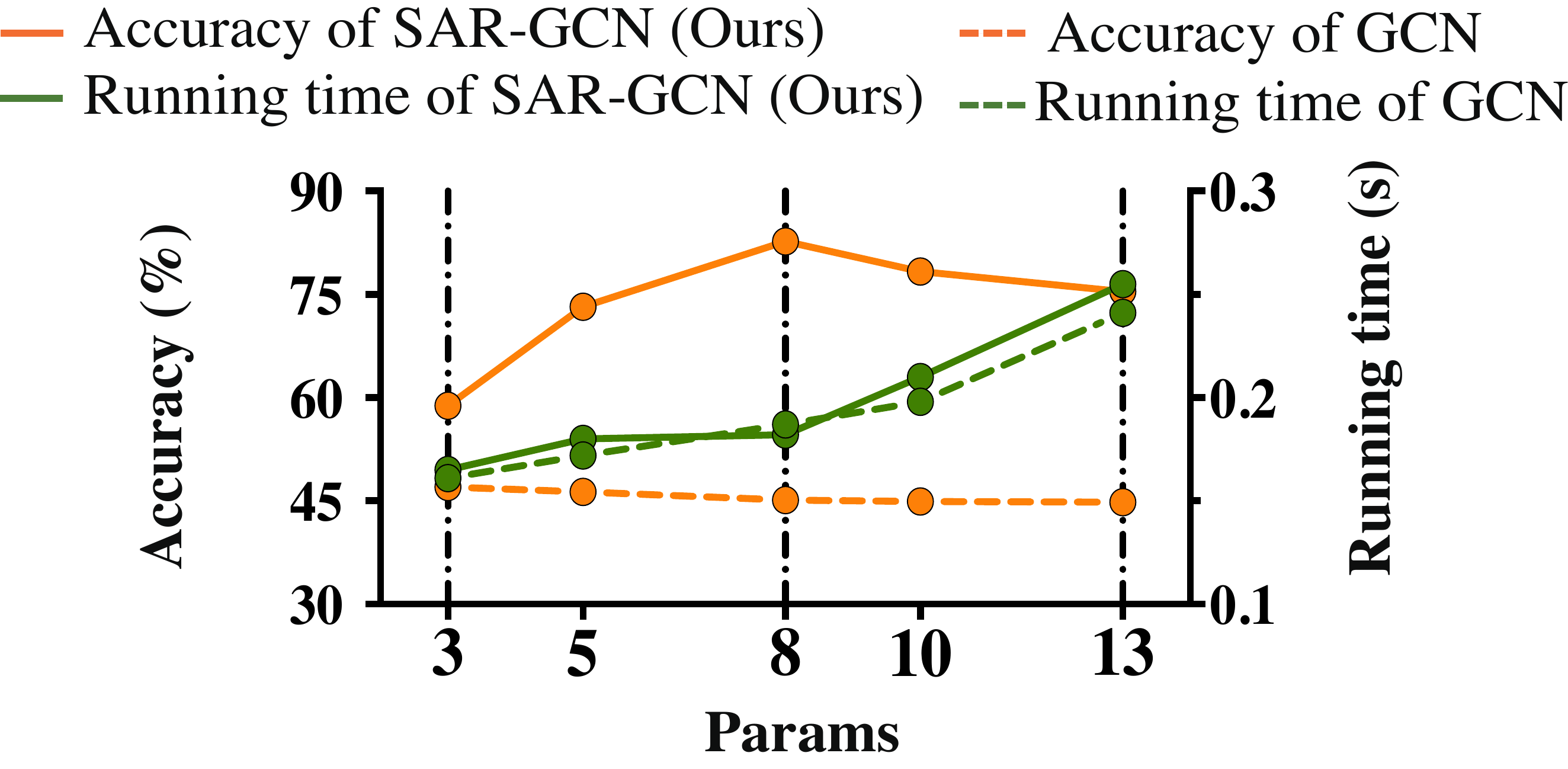}}
\subfigure[] {\includegraphics[width=0.4\textwidth]{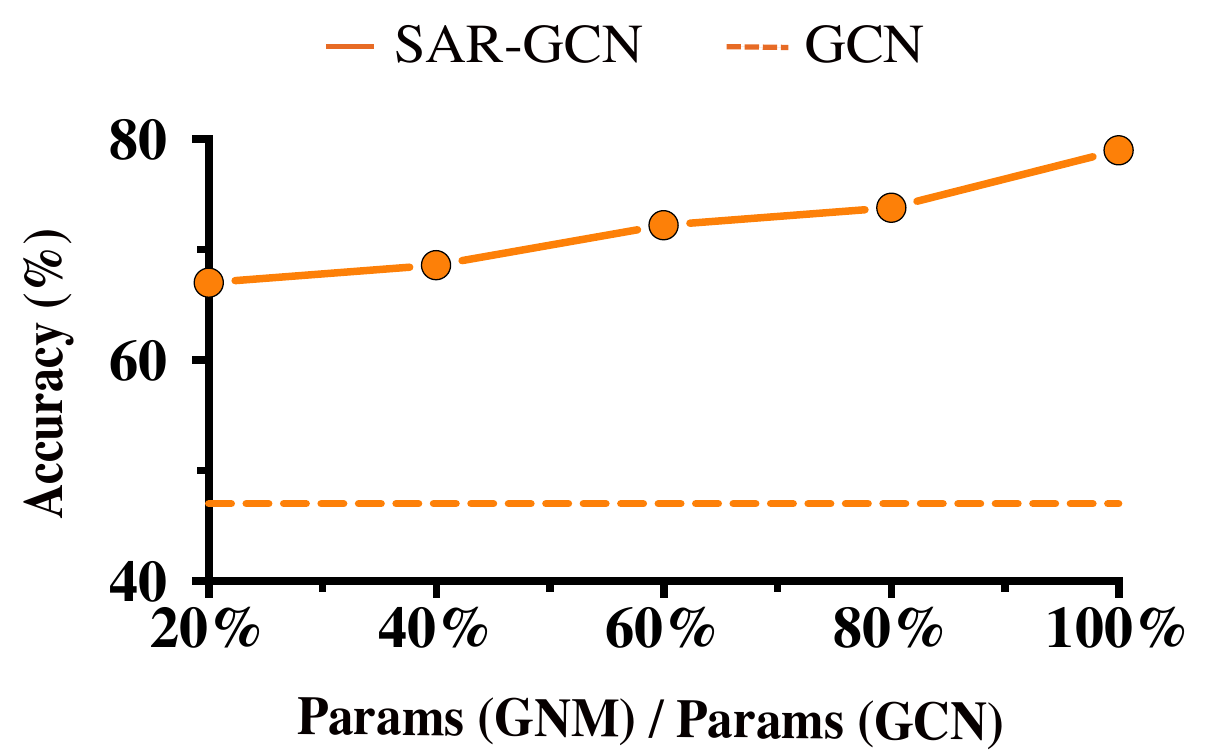}}
\caption{(a) Accuracy and average running time for one graph sample of our SAR-GCN and GCN on TRIANGLES dataset with increasing model size. Note that two models are compared with the same model size in this set of experiments. \MR{(b) Performance of our SAR-GCN on TRIANGLES as a function of increasing model size of the Graph Neural Memory (GNM) built upon a backbone (GCN) with constant model size. The model size of the Graph Neural Memory is expressed as a percentage of the model size of the backbone along the horizontal axis.}}
\label{fig:complexity}
\end{figure}

\smallskip\noindent\textbf{\MR{Model complexity w.r.t. model size.}} \MR{We first conduct experiments to evaluate the model complexity of our model w.r.t. model size.}
\begin{itemize}
    \item \smallskip\noindent\textbf{\MR{Is the performance gain of our model yielded from more learnable parameters compared to the backbone?}} \MR{To investigate it, we compare the performance as well as the running time of both our \emph{SAR-GCN} and the backbone GCN with the same amount of parameters by configuring the model structure of each module. Figure~\ref{fig:complexity} (a) presents the results of both models on \emph{TRIANGLES} dataset as a function of parameter amount, which show that our model consistently outperforms the backbone model significantly with the same amount of parameters, taking comparable running time. Furthermore, our model reaches performance saturation with a larger model capacity than the backbone, revealing the greater potential of our model. Both models exhibit overfitting to some degree when configured with excessive model capacity.}
    
    \item \smallskip\noindent\textbf{\MR{Cost performance in terms of the augmented model size of the Graph Neural Memory.}} \MR{We measure the performance gain of increasing the model size of the Graph Neural Memory with a constant backbone GCN to evaluate the cost performance in terms of the augmented model size between our \emph{SAR-GCN} and the backbone GCN. The results in Figure~\ref{fig:complexity} (b) reveal that our model outperforms the backbone by a large margin even augmented with a relatively small size of Graph Neural Memory. Larger size of Graph Neural Memory yields more performance gain due to more modeling capacity.}
\end{itemize}

\smallskip\noindent\textbf{\MR{Computational complexity.}} \MR{The computational cost of our model mainly consists of 
three parts, namely the cost incurred by Graph Neural Memory, the backbone and the saliency-aware regularization, respectively. We analyze the theoretical computational complexity of these three parts to calculate the overall complexity of our model. We take the instantiation of our model with GCN as an example.}
\begin{itemize}
    \item \MR{\textbf{Complexity of Graph Neural Memory (GNM).} The computational complexity of Graph Neural Memory can be obtained by calculating the cost of each modeling step of $\mathcal{F}_\text{M}$ shown in Equation~\ref{eqn:memory}. Specifically, the complexity for calculating $\mathbf{m}_q$, $\mathbf{K}$ and $\mathbf{V}$ is $O(d_M d)$, $O(N d_H d)$ and $O(N d_H d)$, respectively. Taking into account the complexity involved in the calculation of $\mathbf{M}^l$ and $\mathbf{M}'$, the  complexity for one layer of Graph Neural Memory is: $O(d_M d + N d_H d + N d + d^2)$}.
    
\MR{In practice, $d_M$, $d_H$ and $d$ are approximately in the same order of magnitude, which results in the overall complexity of Graph Neural Memory consisting of $L$ layers:} 
\begin{equation}
        \MR{T_\text{GNM} = O(L k N d^2),}
\end{equation}
\MR{where $k \in \{1, 2\}$ is the iteration number of $\mathcal{F}_\text{M}$ in Equation~\ref{eqn:memory}}. 

    \item \MR{\textbf{Complexity of the saliency-aware regularization.} Measuring the global node saliency formulated in Equation~\ref{eqn:saliency} involves three matrix multiplications, thus the complexity is $O(L d_M d + L N d_H d + L N d)$ which approximately equals to $O(L N d^2)$. Besides, the complexity of regularization with the node saliency, performing weight fusion either by the weighted sum mechanism or the scaling mechanism, is $O(L N^2)$. Thus, the total complexity for the saliency-aware regularization is:}
\begin{equation}
        \MR{T_\text{SAR} = O(L N d^2 + L N^2).}
\end{equation}

    \item \MR{\textbf{Complexity of the backbone (GCN).} The computational complexity of GCN can be derived from the modeling process of GCN in Equation~\ref{eqn:GCN}. It has been analyzed~\cite{Times} that the time complexity of GCN is:}
    \begin{equation}
        \MR{T_\text{GCN} = O(L N d^2 + L|E|d),}
        \label{eqn:complexity_GCN}
    \end{equation}
\MR{where $|E|$ is the number of graph edges.}
\end{itemize}
\MR{Combining the complexity of all three parts, the overall computational complexity of our \emph{SAR-GCN} is:}
\begin{equation}
    \MR{T_\text{SAR-GCN} = O(LNd^2 + LN^2 + L|E|d).}
\end{equation}
\MR{Compared with the complexity of backbone (GCN) in Equation~\ref{eqn:complexity_GCN}, our model has the same order of magnitude of complexity when satisfying $N = O(d^2)$.  }

\begin{figure}[!t]
\centering
\subfigure[] {\includegraphics[width=0.4\textwidth]{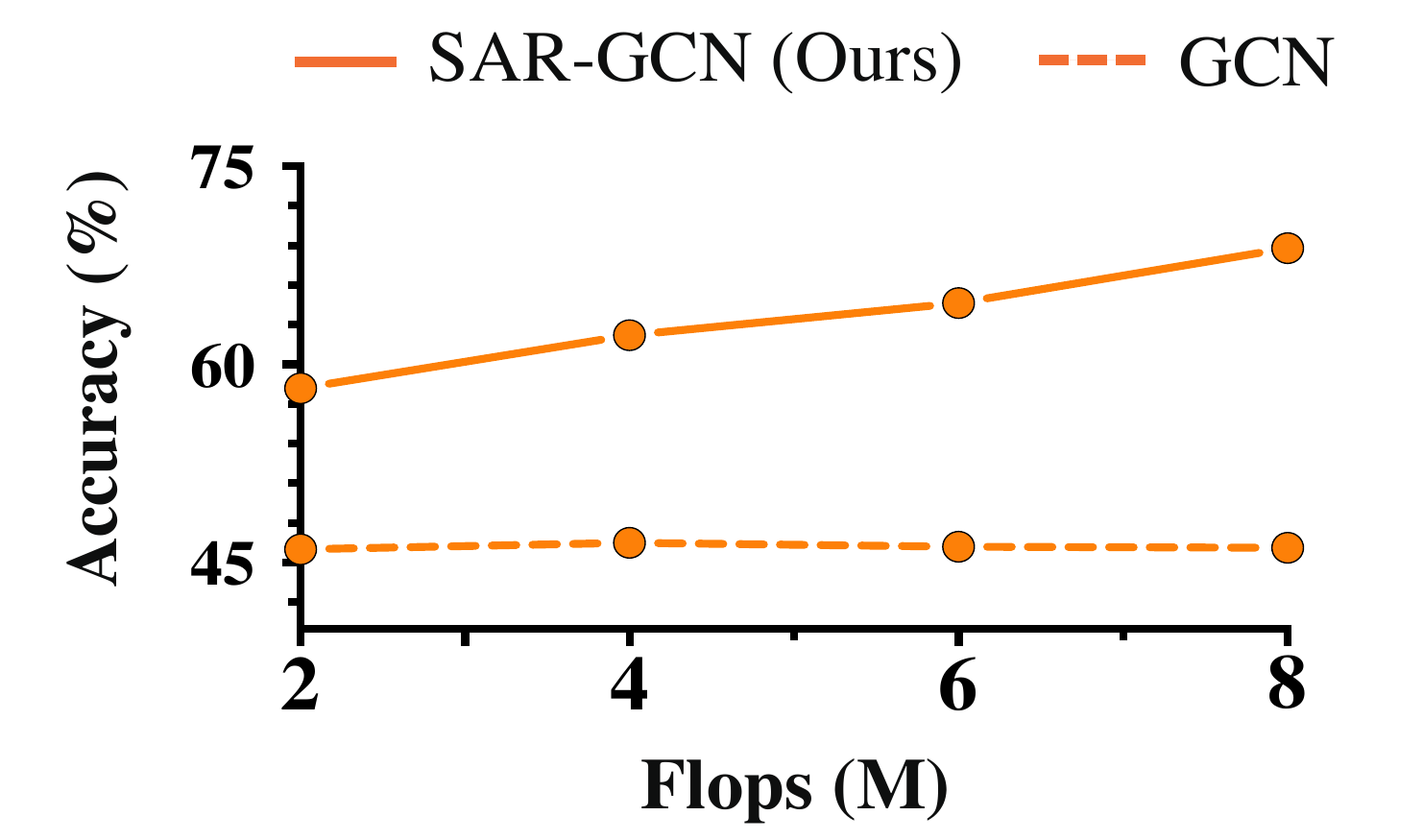}}
\hspace{6mm}
\subfigure[] {\includegraphics[width=0.4\textwidth]{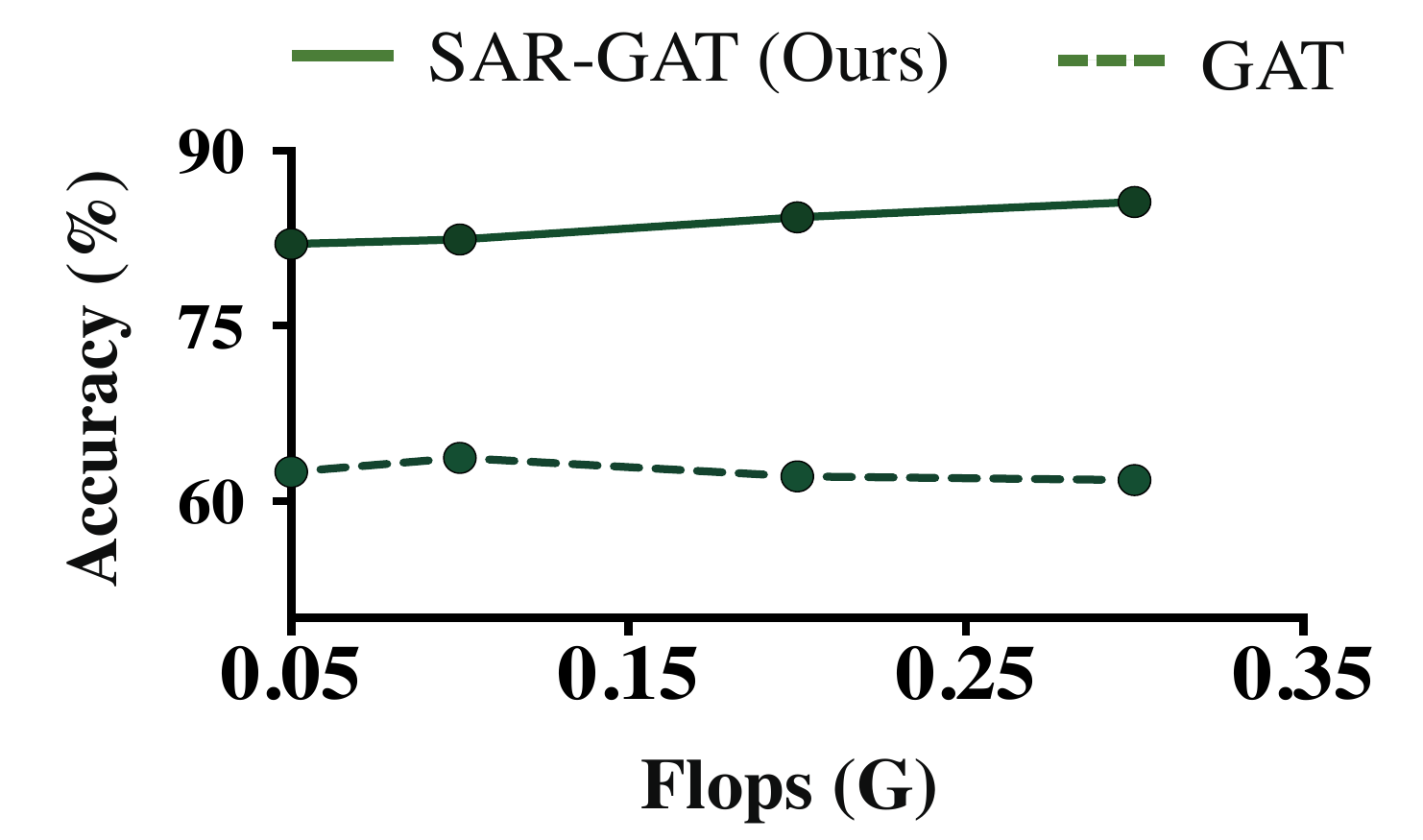}}
\caption{\MR{Accuracy of our model and the corresponding backbone on TRIANGLES dataset as a function of increasing computational complexity. GCN and GAT are used as backbones for comparisons in (a) and (b), respectively.}}
\label{fig:computational_complexity}
\end{figure}

\MR{To have a more comprehensive comparison between our model and the backbone, we conduct experiments to compare the performance of them with the same computational complexity. Figure~\ref{fig:computational_complexity} shows two sets of comparisons using different backbones on TRIANGLES dataset as a function of increasing computational complexity by adjusting the model scale. We observe that our model outperforms the backbones by a large margin consistently, while increasing the model complexity can potentially lead to more superiority. These results demonstrate that the performance gain of our model over the backbone does not result from the extra introduced computational complexity.}




\section{Conclusion}

In this work we have presented the Saliency-Aware Regularized Graph Neural Network (\emph{SAR-GNN}) for graph classification. 
Our \emph{SAR-GNN} learns the global node saliency w.r.t. graph classification, and leverages it to regularize the neighborhood aggregation for feature learning. As a result, our method pays more attention to those nodes that are more relevant to graph classification and is able to learn more effective representations for the entire graph. We have shown the effectiveness of the proposed method built upon different types of graph neural networks by extensive experiments. 






\bibliographystyle{elsarticle-num} 
\bibliography{example_paper}





\end{CJK}
\end{document}